# Sử dụng đa cảm biến cho dẫn đường robot di động tự trị
## *Using multiple sensors for autonomous mobile robot navigation*

**Trần Thuận Hoàng, Phùng Mạnh Dương, Đặng Anh Việt và Trần Quang Vinh**
Trường Đại học Công nghệ, Đại học Quốc gia Hà nội
e-Mail: thuanhoang@donga.edu.vn

**Tóm tắt**
Bài báo trình bày việc sự dụng hệ đo đa cảm biến để dẫn đường cho robot di động tự trị trong nhà. Hệ thống cho phép thu nhận hình ảnh 3D để lập bản đồ toàn cục, và thuật toán giảm thiểu số chiều hình ảnh để thành bản đồ dẫn đường toàn cục 2D, phương pháp thiết kế quỹ đạo bằng phương pháp hàm Lyapunov và tránh vật cản bằng phương pháp trường thế năng cũng được trình bày. Ngoài ra áp dụng phương pháp tổng hợp cảm biến dựa trên bộ lọc Kalman mở rộng cho phép xác định chính xác vị trí và hướng của robot trong điều kiện có can nhiễu của môi trường.

**Abstract**
This paper presents the use of multi-sensor measurement system to guide autonomous mobile robot in the house. The system allows the 3D image acquisition to global mapping, and algorithms to reduce the dimensionality of images to 2D global map navigation, trajectory design approach using the Lyapunov function method and avoid obstacles bythe potential energy can also be presented. Also, sensor integrated method based on extended Kalman filter allows us to identify the exact location and orientation of the robot in the presence of interference from the environment.
**Keyword**: LRF-Laser range finder, 3D-laser image, 2D-laser scanner, nevigation, EKF- extended Kalman filter.

## 1. Nội dung chính

Quá trình dẫn đường cho robot đi động tự trị có thể chia làm 4 bước: lập bản đồ, định vị, tránh vật cản, thiết kế quỹ đạo và điều khiển chuyển động [1]. Với môi trường có cấu trúc, quá trình nhận biết cho phép tạo ra bản đồ hay mô hình không gian phục vụ cho bài toán định vị và thiết kế quỹ đạo robot. Đối với môi trường phi cấu trúc hay thay đổi, robot cần có khả năng tự học quan sát môi trường để xác định được hướng đi của mình. Như vậy, một robot di động tự trị yêu cầu phải định vị chính xác cho hành động dẫn đường của nó. Từ những chỉ số đọc hiện thời của các bộ cảm biến, robot phải xác định chính xác vị trí và định hướng của nó trong môi trường. Có một số loại cảm biến mà có thể được sử dụng cho robot. Mỗi cảm biến đo lường thường chỉ một hoặc hai tham số môi trường với một độ chính xác giới hạn. Tuy nhiên, sử dụng nhiều cảm biến thêm độ chính xác hơn để xác định vị trí của robot. Đó là lý do phương pháp tổng hợp các cảm biến đã được thực hiện trong robot hiện đại để tăng độ chính xác của đo lường. Hầu hết thực hiện phương pháp này được dựa trên suy luận xác suất. Bộ lọc Kalman mở rộng (EKF) là giải pháp xác suất hiệu quả nhất để ước tính đồng thời các vị trí của robot dựa trên một số thông tin về cảm biến bên trong và bên ngoài.

Robot thường sử dụng cảm biến lập mã quang bên trong cho xác định vị trí như một phương pháp đặt tên là odometry [2]. Tuy nhiên, do lỗi tích lũy, sự bất định của vị trí ước tính bởi hệ thống odometric tăng theo thời gian trong khi di chuyển của robot [3]. Để vượt qua những bất lợi này, bằng cách sử dụng một bộ lọc Kalman với các số đo từ một hoặc một số cảm biến bên ngoài kết hợp với đo lường odometric, giá trị vị trí ước tính trở nên chính xác hơn. Điều đó có nghĩa rằng quỹ đạo ước tính là gần hơn với quỹ đạo thật.

Đối với bài toán tránh vật cản trong bài báo này chúng tôi áp dụng phương pháp trường thế năng, với các cảm biến khoảng cách laser và siêu âm, đối với cảm biến laser chúng tôi áp dụng để phát hiện vật cản ở khoảng cách xa robot và cảm biến siêu âm để phát hiện những vật cản gần robot, tuy nhiên các thông tin hình ảnh từ các cảm biến chỉ là thông tin hình ảnh 2D có thể không đủ trong một số trường hợp cần phát hiện các vật có kết cấu không giống nhau theo chiều dọc như cái bàn [4], như biểu diễn trên H.1.

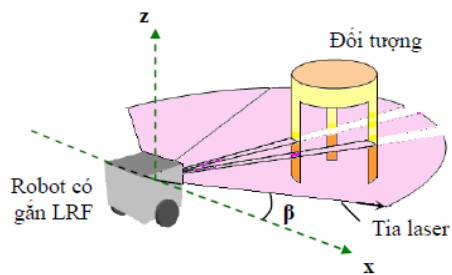

**H. 1**  *Mặt phẳng tia laser với các góc quét β. [4].*

Hình ảnh 2D thu được ở đây chỉ cho phép phát hiện các chân bàn mà không phát hiện được mặt bàn ở vị trí cao hơn và có kích thước lớn hơn nhiều, như vậy áp dụng hệ đo xa 3D cho hình ảnh môi trường bao gồm cả các số đo đối tượng theo chiều dọc nữa. Những thông tin đó sẽ rất cần thiết cho các bài toán tránh vật cản và dẫn đường tối ưu. Tuy nhiên, để dẫn đường cần phải có bản đồ toàn cục 2D, như vậy cần một thuật toán giảm thiểu số chiều hình ảnh từ 3D thành 2D để dẫn đường cho robot.

Cuối cùng, điều khiển hệ ổn định của hệ đã cho theo quỹ đạo mong muốn. Một trong những tiêu chuẩn về tính ổn định được Lyapunov đưa ra hơn một thế kỷ qua vẫn còn nguyên giá trị và ngày càng phát triển. Hai phương pháp do ông đề xuất là phương pháp số mũ đặc trưng và phương pháp hàm Lyapunov. Trong đó, phương pháp hàm Lyapunov được xem là cách tiếp cận chính khi nghiên cứu về tính ổn định, nội dung của phương pháp này là dựa vào sự tồn tại của một lớp hàm đặc biệt (được gọi là hàm Lyapunov) mà tính ổn định của hệ đã cho được kiểm tra trực tiếp qua dấu của đạo hàm (dọc theo quỹ đạo đang xét) của hàm Lyapunov tương ứng.

Do mô hình động học và động lực học của robot di động thường là mô hình phi tuyến dạng hệ nonholonomic (có các ràng buộc về tốc độ chuyển động không khả tích), và có nhiều tham số bất định nên bài toán điều khiển đòi hỏi nhiều nghiên cứu và áp dụng các phương pháp tính toán phức tạp để bảo đảm xe chạy ổn định trơn tru và bám quỹ đạo chính xác. [5].

Nội dung của bài báo được sắp xếp như sau: Mục 2 trình bày tóm tắt xây dựng hình ảnh 3D để lập bản đồ môi trường toàn cục và thuật toán biến đổi ảnh 3D thành 2D phục vụ cho bài toán tránh vật cản và thiết kế quỹ đạo; mục 3 trình bày phương pháp tổng hợp các cảm biến bằng bộ lọc Kalman mở rộng và áp dụng vào mô hình động học của robot để ước lượng chính xác vị trí; mục 4 trình bày phương pháp tránh vật cản bằng phương pháp trường thế; mục 5 thiết kế quỹ đạo và điều khiển chuyển động bằng phương pháp hàm Lyapunov và cuối cùng là các kết quả thực nghiệm với robot được xây dựng và thảo luận.

## 2. Lập bản đồ toàn cục 3D và biến đổi thành bản đồ dẫn đường toàn cục 2D

### 2.1 Kết cấu máy đo xa 3D

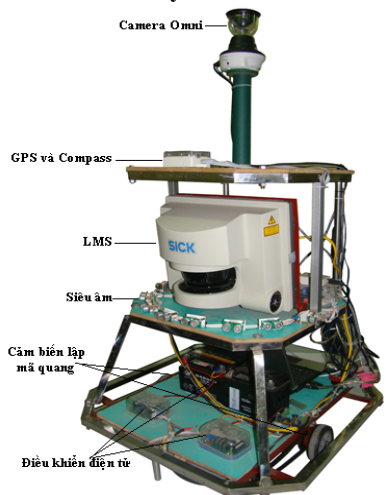

**H. 2**  *Robot Multi-Sensor Smart Robot (MSSR)*

Một máy đo xa laser 2D loại LMS-211 của hãng SICK [5] được sử dụng để xây dựng hệ thống đo 3D-LRF. Do một số vật cản có kết cấu không giống nhau theo chiều dọc như chỉ ra trên H. 2. Hình ảnh 2D thu được ở đây chỉ cho phép phát hiện các chân bàn mà không phát hiện được mặt bàn ở vị trí cao hơn và có kích thước lớn hơn nhiều. Như vậy cần thiết có cảm biến 3D để phát hiện hết toàn bộ chiều cao của vật cản. Với hệ thống chúng tôi, máy LMS-211 hoạt động trong dải đo xa từ cực đại là 8m, góc quét ngang 100° và 180° các điểm đo được quét dần từng bước từ 0° tới 100° hoặc 0° tới 180° với góc tăng dần (độ phân giải) tùy chọn 0,25°; 0,5° hoặc 1°. Máy được gắn lên một đế có thể quay ngẩng lên xuống quanh một trục nằm ngang như H.1 [4].

Hiện tại góc ngẩng đang được đặt: -5°÷30°, và SICK được đặt trên khung robot cao so với mặt đất là h= 40cm, như vậy SICK có thể quét hết một hình 3-D hoàn toàn của một đối tượng ở phía trước với khoảng cách y

$$y = \frac{h}{tg\alpha} \qquad (1)$$

với α là giá trị góc ngẩng; thì robot có thể phát hiện toàn bộ vật cản với chiều cao 200cm ở khoảng cách y = 400 cm như mô tả ở H. 3

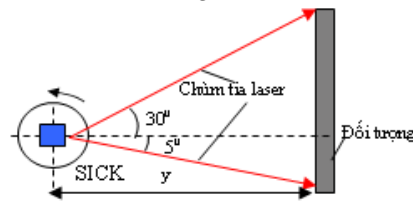

**H. 3**  *Mô tả quét 3D.*

### 2.2. Thuật toán biến đổi ảnh 3D thành 2D cho bài toán dẫn đường

Theo như [4] với thời gian quét 10s ứng với độ phân giải $1^0$, góc quét $100^0$ chúng ta sẽ có tập hợp gồm 81 khung ảnh, mỗi khung ảnh sẽ ứng với chiều cao $z_i$ khác nhau, mỗi tập hợp $z_i$ ta sẽ có 100 điểm $x_i, y_i$ tương ứng với độ phân giải và góc quét hay nói cách khác mỗi khung ảnh sẽ được biểu diễn với tập hợp các giá trị $z_i$ bao gồm: $z_1\{x_1,y_1,…x_{100},y_{100}\},…,z_{81}\{x_1, y_1,…x_{100},y_{100}\}$. Với mục đích phát hiện chiều cao của vật cản mà robot có thể chui qua và tránh, chúng tôi đưa ra giải pháp như sau: chúng tôi chỉ chọn tập hợp các khung ảnh mà ở đó có giá trị $z_i$ chỉ nằm trong giới hạn: 0,8m< $z_i$ <1,2m; (i=1:81), và sau đó đánh dấu tất cả những tập hợp các khung ảnh đó, bước tiếp theo cộng tất cả tập hợp các khung ảnh mà thỏa mãn điều kiện đã nêu, ta sẽ được một mặt phẳng 2D tương ứng có các tọa độ của vật cản mà robot cần phải tránh, các vật cản có chiều cao dưới 0,8m thì có thể được phát hiện bằng cảm biến siêu âm; những vật cản cao hơn 1,2m thì robot có thể chui qua được (chiều cao của robot <1,2m). Thuật toán có thể miêu tả trên H. 4

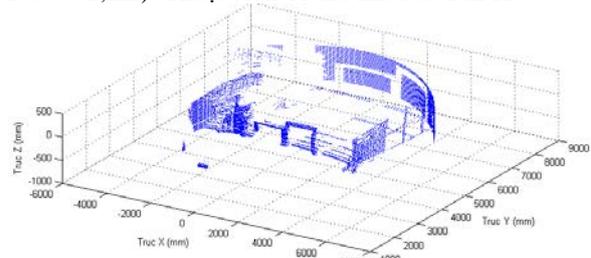

**H. 4**  *Ảnh chụp 3D môi trường toàn cục.*

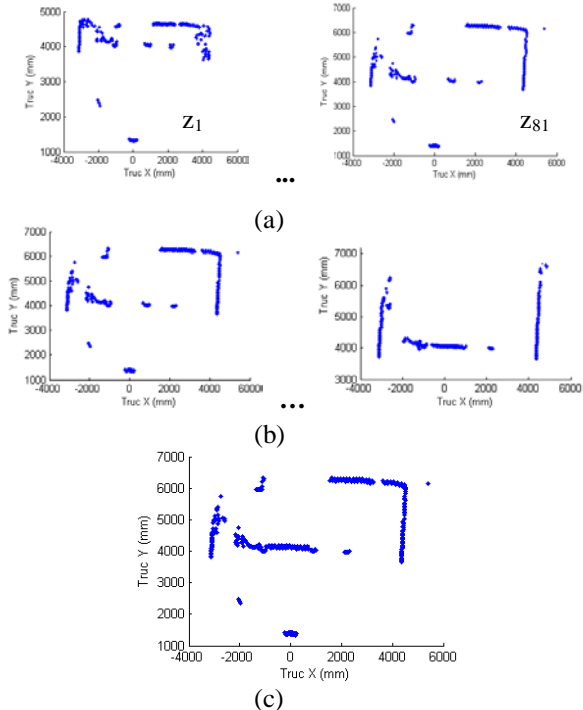

(a)

(b)

(c)

**H. 5** *Thuật toán biến đổi ảnh 3D thành 2D.*
*a)Tập hợp các khung ảnh 2D từ $z_1,...,z_{81}$*
*b)Đánh dấu các khung ảnh thỏa mãn điều kiện*
*c)Cộng các khung ảnh ở (b).*

## 3. Chương trình tổng hợp các cảm biến dùng cho định vị robot di động

Bắt đầu với mô hình động học của robot di động được thiết kế chế tạo của chúng tôi. H. 5 biểu diễn hệ toạ độ và ký hiệu robot, trong đó ($X_G$ $Y_G$) là hệ tọa độ toàn cục, ($X_R$, $Y_R$) là tọa độ cục bộ gắn với tâm robot. R là bán kính của bánh xe, và L là khoảng cách giữa các bánh xe.

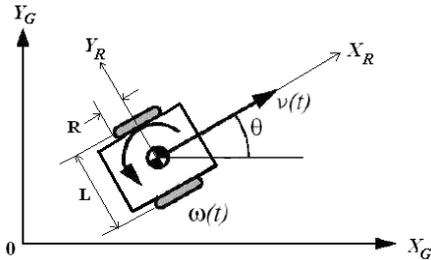

**H. 6** *Thế và các tham số của robot.*

Trong khoảng thời gian lấy mẫu số liệu đo $\Delta t$, các tốc độ góc của các bánh xe bên trái $\omega_L$ và bên phải $\omega_R$ tạo ra các lượng di chuyển tương ứng của bánh trái $\Delta s_L$ và bánh phải $\Delta s_R$:

$$\Delta s_L = \Delta t.R. \omega_L \; ; \; \Delta s_R = \Delta t.R. \omega_R$$

Từ đó dẫn đến sự dịch chuyển của tâm robot $\Delta s$ và góc hướng của robot $\Delta \theta$:

$$\Delta s = \frac{\Delta s_L + \Delta s_R}{2} \; ; \; \Delta \theta = \frac{\Delta s_R - \Delta s_L}{L}$$

Trạng thái của robot tại thời điểm k+1 trong khung tọa độ toàn cục có thể được cập nhật như sau:

$$\begin{bmatrix} x_{k+1} \\ y_{k+1} \\ \theta_{k+1} \end{bmatrix} = \begin{bmatrix} x_k \\ y_k \\ \theta_k \end{bmatrix} + \begin{bmatrix} \Delta s_k \cos(\theta_k + \Delta \theta_k / 2) \\ \Delta s_k \sin(\theta_k + \Delta \theta_k / 2) \\ \Delta \theta_k \end{bmatrix} \quad (2)$$

Trong thực tế, hệ thống (2) gặp phải các sai số hệ thống. Có nhiều công trình cố gắng tăng độ tin cậy của việc định vị này trong đó sử dụng bộ lọc Kalman. Chúng tôi cũng đã áp dụng phương pháp tổng hợp cảm biến bằng thuật toán lọc Kalman để định vị cho robot được chế tạo và đã nhận được kết quả cải thiện được chất lượng của các phép đo này đáng kể.

Có thể tóm tắt phương pháp này như sau. Nếu $X = [x \; y \; \theta]^T$ là vectơ trạng thái của robot đang ở tọa độ $x$, $y$ và góc hướng $\theta$, trên mặt phẳng mà nó di chuyển, trạng thái này có thể được quan sát bởi một số phép đo tuyệt đối, $z$. Các phép đo này được miêu tả bởi một hàm phi tuyến, $h$, của hệ toạ độ robot và một quá trình nhiễu Gaussian độc lập, $v$. Tổng quát, (3) là một hàm phi tuyến, $f$, với một vector đầu vào $u$, có phương trình trạng thái và phương trình lối ra như sau:

$$x_{k+1} = f(x_k, u_k) + w_k \quad (3)$$
$$z_k = h(x_k) + v_k \quad (4)$$

Trong đó các biến ngẫu nhiên $w_k$ và $v_k$ biểu diễn cho quá trình và nhiễu đo tương ứng. Chúng được giả định là độc lập với nhau, ồn trắng, và với phân bố xác suất chuẩn:

$$w_k \sim N(0,Q_k) \quad v_k \sim N(0,R_k) \quad E(w_i v_j^T) = 0$$

Dựa trên các số liệu đo $z$, có thể tìm được một hệ số Kalman $K$ trong mỗi chu kỳ lấy mẫu tín hiệu đo thuộc một vòng lặp đệ quy gọi là bộ lọc Kalman, sao cho giá trị ước lượng trạng thái của hệ gần với giá trị thực nhất. Bộ lọc Kalman mở rộng, áp dụng cho các hệ phi tuyến, được thực hiện qua các bước như sau [11]:

1. *Bước dự báo* với các phương trình cập nhật:

$$\hat{x}_k^- = f(\hat{x}_{k-1}, u_{k-1}) \quad (5)$$
$$P_k^- = A_k P_{k-1} A_k^T + W_k Q_{k-1} W_k^T \quad (6)$$

Trong đó $\hat{x}_k^- \in \Re^n$ là ước lượng trạng thái tiên nghiệm tại thời điểm k cho biết giá trị trước quá trình tại thời điểm k.

$\hat{P}_k^-$ là ma trận hiệp phương sai của sai số dự báo trạng thái.

$A_k$ là Jacobi của đạo hàm riêng của $f$ theo $x$.

$W_k$ là Jacobi của đạo hàm riêng của $f$ theo $w$.

$Q_{k-1}$ là ma trận hiệp phương sai nhiễu đầu vào phụ thuộc vào độ lệch chuẩn của nhiễu của tốc độ góc của các bánh xe. Chúng được mô hình hóa như là tỷ lệ thuận với tốc độ góc $\omega_{R-k}$, và $\omega_{L-k}$ của các bánh xe tại thời điểm k. Điều này dẫn đến phương sai bằng $\delta\omega_R^2$ và $\delta\omega_L^2$, trong đó $\delta$ là một hằng số xác định bởi các thực nghiệm.

Ma trận hiệp phương sai Q được xác định:

$$Q_k = \begin{bmatrix} \delta.\omega_{R,k}^2 & 0 \\ 0 & \delta.\omega_{L,k}^2 \end{bmatrix} \quad (7)$$

2. *Bước hiệu chỉnh* với các phương trình cập nhật phép đo:

$$K_k = P_k^- H_k^T (H_k P_k^- H_k^T + V_k R_k V_k^T)^{-1} \quad (8)$$

$$\hat{x}_k = \hat{x}_k^- + K_k (z_k - h(x_k^-, 0)) \quad (9)$$

$$P_k = (I - K_k H_k) P_k^- \quad (10)$$

Trong đó $\hat{x}_k \in \Re^n$ là ước lượng trạng thái hậu nghiệm tại thời điểm k, giá trị thu được sau khi đo lường $z_k$.

$K_k$ là hệ số độ lợi Kalman.

$V_k, H_k$ là các Jacobi đạo hàm riêng của $h$ theo $v$.

$R_k$ là ma trận hiệp phương sai của nhiễu được ước lượng từ nhiễu trong phép đo của bộ mã hóa bánh xe và các cảm biến khác. Các phép đo của các bộ cảm biến này được thu thập vào trong véc tơ $z_k$ như sau:

Bằng phương trình hệ thống (3), các tham số trạng thái ($x_{m\cdot odometry}, y_{m\cdot odometry}$ và $\theta_{m\cdot odometry}$) nhận gián tiếp bằng các chỉ số đọc từ bộ mã hóa bánh xe.

Góc hướng robot có thể cũng được đo trực tiếp bởi phép đo tuyệt đối với cảm biến từ-địa bàn $\theta_{m.compass}$.

Kết hợp các phép đo gián tiếp từ Odometry và các phép đo trực tiếp từ các cảm biến như nói trên ta có ma trận $z_k$ trong phương trình sửa giá trị ước lượng (9) có dạng như sau:

$$z_k = \begin{pmatrix} x_{odometry} & y_{odometry} & \theta_{odometry} & \theta_{compass} & r_{LRF} & \psi_{LRF} & \varphi_{camera} \end{pmatrix}^T \quad (11)$$

Ma trận hiệp phương sai $R_k$ của nhiễu đo có cấu trúc đường chéo. Nhiễu của phép đo tốc độ bánh xe có thể được xác định bởi thực nghiệm. Sự chính xác của các cảm biến từ-địa bàn, của các phép đo LRF và cameara nhận được từ đặc điểm kỹ thuật của nhà sản xuất. Các số liệu này được điền vào $R_k$ cho bước điều chỉnh EKF.

$$R_k = \begin{bmatrix} \text{var}(\omega_L) & 0 & 0 & 0 & 0 & 0 \\ 0 & \text{var}(\omega_R) & 0 & 0 & 0 & 0 \\ 0 & 0 & \text{var}(\theta_{compass}) & 0 & 0 & 0 \\ 0 & 0 & 0 & \text{var}(r_{LRF}) & 0 & 0 \\ 0 & 0 & 0 & 0 & \text{var}(\psi_{LRF}) & 0 \\ 0 & 0 & 0 & 0 & 0 & \text{var}(\varphi_{camera}) \end{bmatrix} \quad (12)$$

## 4. Tránh vật cản bằng phương pháp trường thế

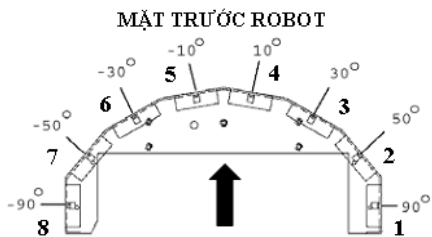

**H. 7** *Sơ đồ bố trí cảm biến siêu âm*

Chúng tôi bố trí lắp đặt siêu âm như trên H.7, do chỉ có các thiết bị siêu âm phía trước và hai bên robot nên ta chỉ xét ảnh hưởng của các vật cản ở phía trước và hai bên robot.

Gọi tọa độ của mục tiêu là $q_{goal} = (X_{goal}, Y_{goal})$, tọa độ của robot là $q = (X, Y)$. Lực tạo ra từ mục tiêu $F_{att}(q) = -\nabla U_{att}(q) = -k_{att}(q - q_{goal})$ tạo ra vận tốc dài và vận tốc góc hướng về mục tiêu là:

$$V_{att}(q) = -k_{att}\left[(X - X_{goal})\cos(\theta_{goal}) + (Y - Y_{goal})\sin(\theta_{goal})\right] \quad (13)$$

$$\omega = -k_{att}(\theta - \theta_{goal}) \quad (14)$$

Với $\theta_{goal} = \text{atan2}(Y_{goal} - Y, X_{goal} - X)$ là góc chỉ hướng của đường thẳng nối từ robot tới mục tiêu; $\theta$ là góc chỉ hướng tức thời của robot trong hệ tọa độ Đề Các OXY.

Khi robot đi vào vùng lân cận (khoảng cách tới vật cản nhỏ hơn $d_0$) vật cản thứ $i$ và vật cản nằm trong phạm vi $(\theta - 60^o, \theta + 60^o)$ trong hệ tọa độ cực có gốc tọa độ là tâm của robot, thì vận tốc dài và vận tốc góc robot là:

$$V_{rep}^i(q) = \begin{cases} \dfrac{1}{2} k_{obst}^i \left(\dfrac{1}{d_{obst}^i(q)} - \dfrac{1}{d_0}\right) \dfrac{1}{\left(d_{obst}^i(q)\right)^2} & d_{obst}^i(q) < d_0 \\ 0 & d_{obst}^i(q) \geq d_0 \end{cases} \quad (15)$$

$$\omega = -k_{att}^i(\theta - \theta_{obs}^i) \quad (16)$$

Trong đó: nếu vật cản $obs_i$ ở phía trước, bên phải robot thì $\theta_{obs}^i = \text{atan2}(Y - Y_{goal}, X - X_{goal}) + 60^o$

nếu vật cản $obs_i$ ở phía trước, bên trái robot thì $\theta_{obs}^i = \text{atan2}(Y - Y_{goal}, X - X_{goal}) - 60^o$

Trong ứng dụng này, ta lựa chọn các tham số như sau: $d_0 = 0,7m \quad k_{att} = 10 \quad k_{rep}^i = 10$

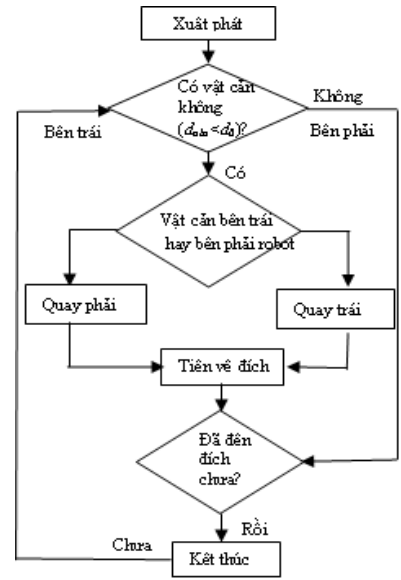

**H. 8** *Lưu đồ thuật toán tránh vật cản bằng phương pháp trường thế.*

# 5. Thiết kế quỹ đạo và điều khiển chuyển động bằng phương pháp hàm Lyapunov

Để mà dẫn đường cho robot đi đến đích ngoài những điều kiện ràng buộc về điều kiện nonholonomic của hệ thống bánh xe mà còn đáp ứng điều kiện tối ưu của luật điều khiển [5][6][7]

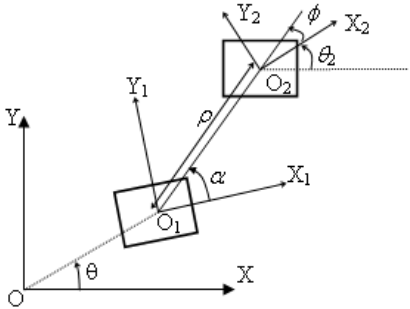

**H. 9** *Các thông số và thế của robot* [5]

Từ H.9 luật điều khiển có thể được tóm tắt như sau: Với 2 điểm tham chiếu trong mặt phẳng OXY là $O_1$ và $O_2$ mà robot đi qua, thế của robot bao gồm vị trí và hướng của robot tại $O_1$ là $X_1, Y_1, \theta_1$ và $O_2$ là $X_2, Y_2, \theta_2$. Với $\rho$ là khoảng cách từ $O_1$ đến $O_2$; $\phi$ là góc tạo bởi hai véc tơ $O_1O_2$ với $O_2X_2$; $\alpha$ là góc tạo bởi hai véc tơ $O_1O_2$ với $O_1X_1$.

$$\begin{cases} \rho = \sqrt{(X_2-X)^2 + (Y_2-Y)^2} \\ \phi = \operatorname{atan2}(Y_2-Y, X_2-X) - \theta_2 \\ \alpha = \operatorname{atan2}(Y_2-Y, X_2-X) - \theta \end{cases} \quad (17)$$

Luật điều khiển ràng buộc vận tốc để đảm bảo quỹ đạo của robot đi từ điểm ban đầu $O_1$ đến đích $O_2$ với các biến giá trị điều khiển ($\rho, \alpha, \phi$), được gọi là các biến dẫn đường, và chúng sẽ hội tụ về zero khi robot tiến về đích.

Mô hình động học của robot di động có thể được biểu diễn như sau:

$$\begin{cases} \dot{X} = v\cos\theta \\ \dot{Y} = v\sin\theta \\ \dot{\theta} = \omega \end{cases} \quad (18)$$

Trong đó, $\omega$ và $v$ lần lượt là vận tốc góc và vận tốc dài của robot. Mô hình động học của robot được mô tả qua các biến navigation ($\rho, \alpha, \phi$) như sau:

$$\begin{cases} \dot{\rho} = -v\cos\alpha \\ \dot{\alpha} = -\omega + v\sin\alpha/\rho \\ \dot{\phi} = v\sin\alpha/\rho \end{cases} \quad (19)$$

Từ [6] hàm lyapunov được xây dựng như sau:

$$V = V_{g1} + V_{g2} = \rho^2/2 + (\alpha^2 + h\phi^2)/2$$
$$v = k_v \rho \cos\alpha \quad (20)$$
$$\omega = k_\alpha \alpha + (k_v \cos\alpha \sin\alpha/\alpha)(\alpha + h\phi)/\rho$$

Trong đó $k_v$ và $k_\alpha$ là các hệ số vận tốc, theo [6] để đáp ứng điều kiện tối ưu của luật điều khiển thì đạo hàm bậc nhất của $V_{g1}$ và $V_{g2}$ luôn luôn có giá trị âm do đó các biến dẫn đường sẽ hội tụ về zero ở tại tọa độ đích.

Gọi tọa độ thật của robot là $(\hat{X}, \hat{Y}, \hat{\theta})$, tọa độ đo được của robot là $(X, Y, \theta)$.

Đặt $\varepsilon_X, \varepsilon_Y, \varepsilon_\theta$ là nhiễu đo của phép đo vị trí $(\hat{X}, \hat{Y}, \hat{\theta})$ của robot. Các giá trị đo của vị trí và hướng X, Y, θ như sau:

$\hat{X} = X + \varepsilon_X$, $\hat{Y} = Y + \varepsilon_Y$, $\hat{\theta} = \theta + \varepsilon_\theta$. Trong đó $|\varepsilon_X| \leq \|\varepsilon_X\|$, $|\varepsilon_Y| \leq \|\varepsilon_Y\|$, $|\varepsilon_\theta| \leq \|\varepsilon_\theta\|$. Đặt $\varepsilon_\rho, \varepsilon_\alpha, \varepsilon_\phi$ lần lượt là các thành phần nhiễu của các biến navigation:

$$\varepsilon_\rho = \sqrt{(X_2-\hat{X})^2 + (Y_2-\hat{Y})^2} - \sqrt{(X_2-X)^2 + (Y_2-Y)^2}$$
$$\varepsilon_\phi = \sqrt{(X_2-\hat{X})^2 + (Y_2-\hat{Y})^2} - \sqrt{(X_2-X)^2 + (Y_2-Y)^2} \quad (21)$$
$$\varepsilon_\alpha = \varepsilon_\phi - \varepsilon_\theta$$

Giá trị thật của các biến navigation là:

$$\begin{cases} \hat{\rho} = \sqrt{(X_2-\hat{X})^2 + (Y_2-\hat{Y})^2} \\ \hat{\phi} = \operatorname{atan2}(Y_2-\hat{Y}, X_2-\hat{X}) - \theta_2 \\ \hat{\alpha} = \operatorname{atan2}(Y_2-\hat{Y}, X_2-\hat{X}) - \hat{\theta} \end{cases} \quad (22)$$

để xác định các yếu tố đầu vào $\omega$ và $v$ để điều khiển robot đi theo quỹ đạo (4). Các nhiễu đo ($\varepsilon_X, \varepsilon_Y, \varepsilon_\theta$) ảnh hưởng đáng kể đến hiệu quả của mô hình điều khiển, bộ lọc Kalman mở rộng được sử dụng để tổng hợp dữ liệu từ các cảm biến và chuyển động học để ước lượng chính xác hơn trạng thái của hệ thống như biểu diễn trên H.10

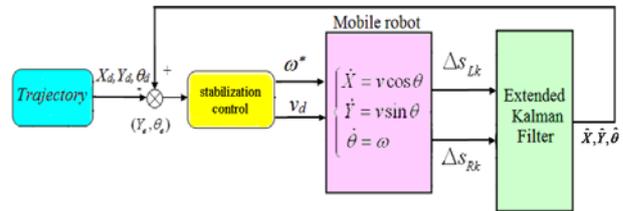

**H. 10** *Mô hình điều khiển của thiết kế quỹ đạo* [5]

## 6. Thực nghiệm và thảo luận

Robot đã được thiết kế, chế tạo và cho chạy thử nghiệm trong một phòng có mặt sàn gỗ phẳng [5]. H.11 cho thấy sự cải thiện đáng kể khi áp dụng phương pháp tổng hợp cảm biến.

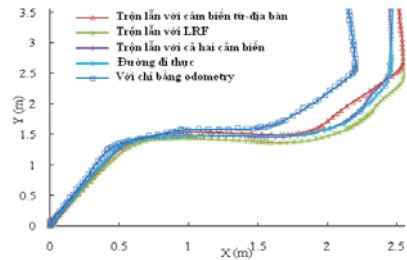

**H. 11** *Quỹ đạo robot ước lượng được với các cấu hình bộ lọc EFK khác nhau* [5]

Với một thí nghiệm khác, chúng tôi đánh giá khả năng có thể sử dụng được của bộ điều khiển đã đề xuất với áp dụng cho robot tự trị. Mục đích để dẫn đường robot bắt đầu từ điểm khởi đầu có tọa độ(X,Y,θ) là (0,0,0) đi đến các điểm có tọa độ $(2,2,30^0)$, $(2,2,60^0)$ và $(2,2,90^0)$. H. 12 biểu diễn sự thay đổi vận tốc dài và vận tốc góc của robot

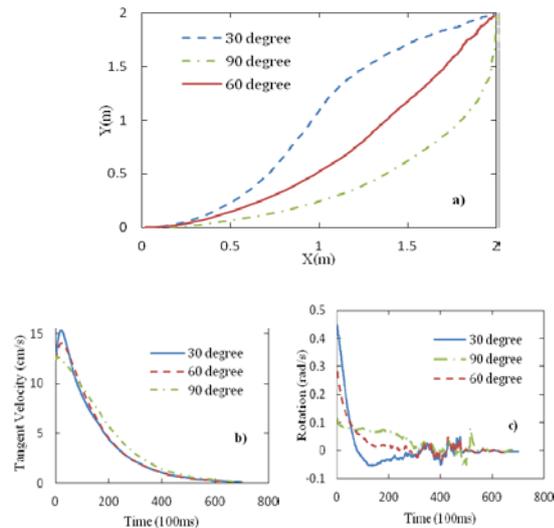

**H. 12** *Kết quả của điều khiển ổn định*
*a) Quỹ đạo của robot; b)Vận tốc dài;c) Vận tốc góc*

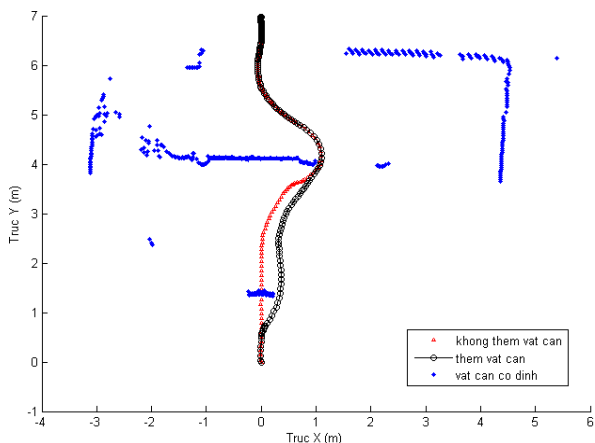

**H. 13** *Quỹ đạo tránh vật cản của robot*

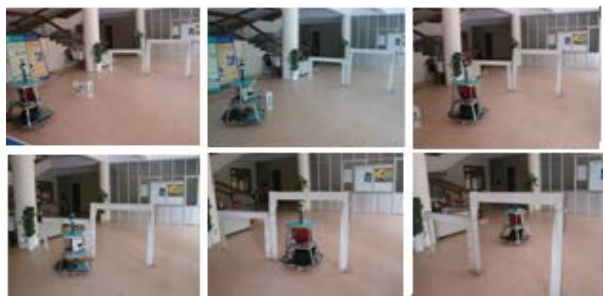

**H. 14** *Ảnh chụp liên tiếp chuyển động tránh vật cản của robot*

Và một thí nghiệm cho dẫn đường robot di động bằng bản đồ toàn cục 2D mà được xây dựng từ bản đồ toàn cục 3D. Chúng tôi sẽ cho robot đi thẳng với tọa độ (0,0,0) và tọa độ đích là (0,7,0), robot sẽ đi theo quỹ đạo theo thuật toán đã trình bày ở trên và dùng bản đồ dẫn đường 2D để tránh vật cản, trong quá trình di chuyển nếu gặp những vật cản bất ngờ thì robot sẽ tránh vật cản bằng phương pháp trường thế.

Trên H.13 chỉ ra đường quỹ đạo mà robot phải tránh vật cản, đường quỹ đạo thứ nhất robot tránh một cổng chào có độ cao thấp hơn robot nhờ hình ảnh 3D ở H.4 và thuật toán H.5; đường quỹ đạo thứ 2 trên đường đi như đường quỹ đạo thứ nhất phát hiện thêm một vật cản và tự tránh nó bằng phương pháp trường thế; H.14 là ảnh chụp liên tiếp của robot tránh vật cản của đường quỹ đạo thứ 2.

## 7. Kết luận

Trong báo cáo này, chúng tôi đã phát triển một hệ thống robot đa cảm biến cho phép thu nhận hình ảnh 3D để lập bản đồ toàn cục và dẫn đường tự động. Chúng tôi cũng đề xuất thuật toán giảm thiểu số chiều hình ảnh để thành bản đồ dẫn đường toàn cục 2D, phương pháp thiết kế quỹ đạo bằng phương pháp hàm Lyapunov và tránh vật cản bằng phương pháp trường thế năng cũng được trình bày. Kỹ thuật tổng hợp cảm biến dựa trên bộ lọc Kalman mở rộng cũng đã được triển khai. Nhiều phép đo đạc thực nghiệm đã được tiến hành chứng minh hiệu quả của hệ thống đề xuất.